\documentclass[letterpaper, 10 pt, conference]{ieeeconf}  

\IEEEoverridecommandlockouts                              

\usepackage{graphics} 
\usepackage{epsfig} 
\usepackage{times} 
\usepackage{amsmath} 
\usepackage{amssymb}  
\usepackage{mathtools}
\usepackage{subcaption}
\usepackage{booktabs}
\usepackage{adjustbox}
\usepackage{caption}
\usepackage{stfloats}
\usepackage[colorlinks=true, allcolors=blue]{hyperref}

\title{\LARGE \bf \textit{FRAME:} Fast and Robust Autonomous 3D point cloud Map-merging for Egocentric multi-robot exploration}

\author{Nikolaos Stathoulopoulos$^1$, Anton Koval$^1$, Ali-akbar Agha-mohammadi$^2$ and George Nikolakopoulos$^1$
\thanks{This work has been funded by the European Unions Horizon 2020 Research and Innovation Programme under the Grant Agreement No. 101003591 NEX-GEN SIMS.}
\thanks{$^{1}$The Authors are with the Robotics and AI Group, Department of Computer, Electrical and Space Engineering, Lule\r{a} University of Technology, 971 87 Lule\r{a}, Sweden} %
\thanks{$^{2}$The author is with Jet Propulsion Laboratory California Institute of Technology Pasadena, CA, 91109.}
\thanks{Corresponding Author's Email: \texttt{niksta@ltu.se}}}%
\begin{document}
\maketitle
\thispagestyle{empty}
\pagestyle{empty}


\begin{abstract}
This article presents a 3D point cloud map-merging framework for egocentric heterogeneous multi-robot exploration, based on overlap detection and alignment, that is independent of a manual initial guess or prior knowledge of the robots' poses.
The novel proposed solution utilizes state-of-the-art place recognition learned descriptors, that through the framework's main pipeline, offer a fast and robust region overlap estimation, hence eliminating the need for the time-consuming global feature extraction and feature matching process that is typically used in 3D map integration. The region overlap estimation provides a homogeneous rigid transform that is applied as an initial condition in the point cloud registration algorithm Fast-GICP, which provides the final and refined alignment.
The efficacy of the proposed framework is experimentally evaluated based on multiple field multi-robot exploration missions in underground environments, where both ground and aerial robots are deployed, with different sensor configurations.
\end{abstract}


\section{INTRODUCTION}

In recent years, great emphasis has been given to researching and developing techniques for locating and reconstructing unknown environments autonomously, with numerous real-world applications, such as mine exploration~\cite{dark_mine}, planetary exploration~\cite{planetary}, search and rescue missions~\cite{compra}, industrial inspection~\cite{ind_insp} and so on. However, the necessity to accommodate for greater and more complex environments, in terms of time and efficiency, has stimulated an extensive research trend in Multi-Robot Systems (MRS)~\cite{nebula}.
Thus, these systems aim at obtaining a consistent and robust method that allows any number of agents to cooperate, explore and map an environment, while being more time efficient, through task parallelization, as well as providing a higher level of reliability, redundancy and resiliency. 

It is more than evident that all the aforementioned applications share the need for an autonomous map-merging procedure, especially when multi-robot systems are deployed in the field. In this case, each robot generates a local map, within its local frame, that typically serves as a source of information for localization, collision avoidance, navigation and path planning, and can later on be shared and fused into a global map. The complexity of the map-merging process is determined by a variety of factors, including the level of awareness about the agents' relative positions and orientations, as well as the correct data fusion of sensors equipped on each robot, which may have varying degrees of accuracy, inherent noise or range. 
Multi-modal systems are common during long exploration missions~\cite{multi_modal}, while variances in sensors of different robotic platforms, aerial or ground, can cause the individual maps to vary substantially and thus make the direct fusion of the maps more difficult. 

\begin{figure}[b!]
    \centering
    \vspace{-5mm}
    \includegraphics[width=1.\columnwidth]{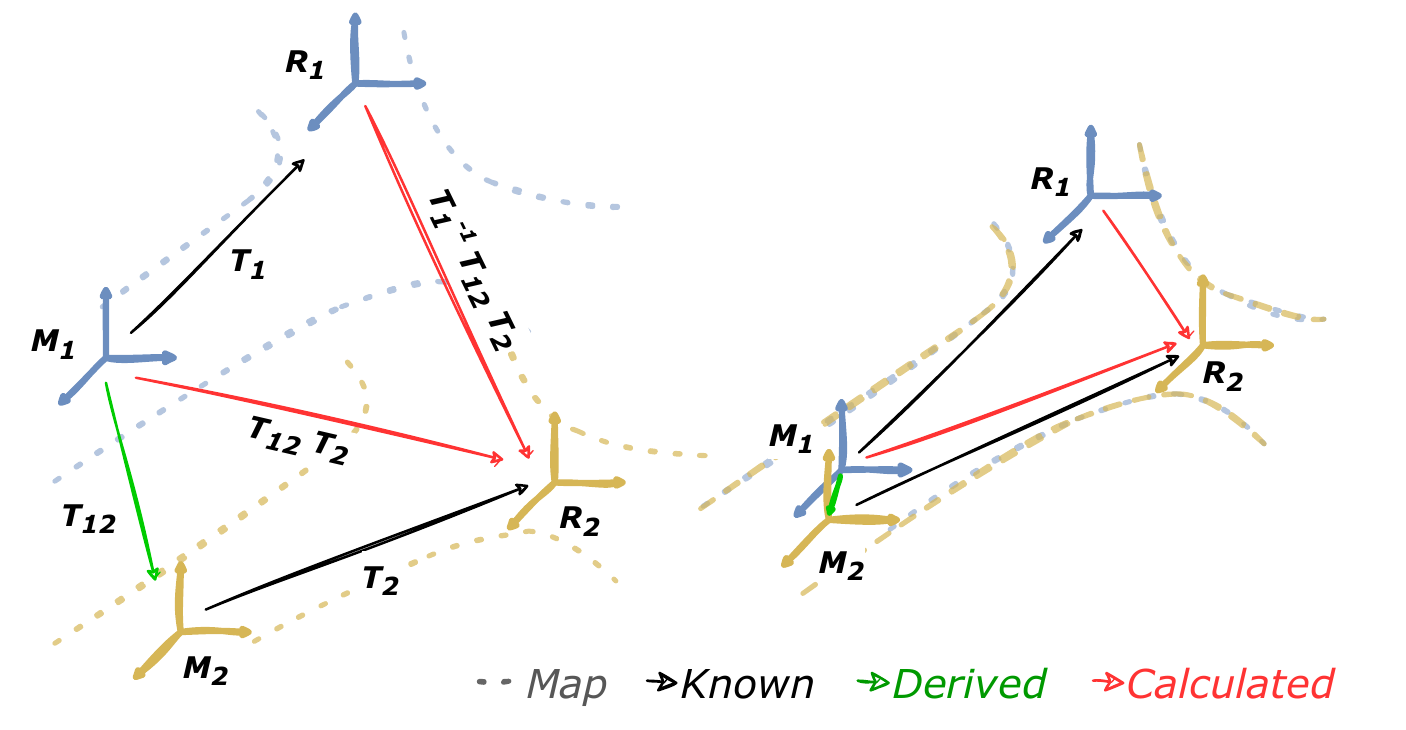}
    \setlength{\abovecaptionskip}{-10pt}
    \caption{The map $\mathcal{M}_1$, $\mathcal{M}_2$ and the robot $\mathcal{R}_{1}$, $\mathcal{R}_{2}$ frames, before and after the map alignment. $T_1$ and $T_2$ are the non-static known transforms, $T_{12}$ is the spatial coordinate transform derived from our framework, while the rest of the transforms between each robot and each map can be calculated from the aforementioned.}
    \label{fig:frames}
\end{figure}

In this article, we introduce a novel framework for Fast and Robust Autonomous 3D point cloud Map-merging for Egocentric multi-robot exploration, which we will refer to as FRAME, tackling the problem of identifying acceptable spatial coordinate transformations between local maps, also known as map alignment. 
The result, as depicted in Fig.~\ref{fig:frames}, can also determine the relative robot poses, which for many existing map alignment methods, is a limiting prerequisite. Other assumptions, present in the current literature, are the identical map formats, the need for an initial guess and the high map overlap ratio~\cite{alignment, planar_features}. 
In contrary, in our proposed approach, we only make the assumption of a partial overlap, which is detected through learned, place dependent descriptors, thus eliminating the need for the time costly global feature extraction process. 
Along with the learned orientation regression descriptors, we yield an initial homogeneous rigid transform that is later used for a faster and more accurate convergence of the registration algorithm that refines the alignment between the two point cloud maps.


\subsection*{— Contributions}

Therefore, our contributions can be summarized as follows: (a) Unlike most solutions, found in the current literature that rely on 2D grid maps, we introduce a 3D point cloud map-merging framework that offers low enough processing times that can be used in real-time multi-robot exploration missions, increasing the mapping efficiency, since repeated exploration of the same places can be avoided. (b) The proposed framework does not require a manual initial guess, hence promoting robotic autonomy by decreasing the need for human intervention. (c) We present a novel application of place-recognition deep-learned descriptors, replacing the time-consuming global feature extraction, that is mainly performed with handcrafted descriptors in the problem of map-merging. 
The evaluation is experimentally verified in harsh field environments with both aerial and ground robots, while using different sensors and mapping configurations, offering a robust solution to multi-modal systems.

\section{RELATED WORK} \label{sec:related_work}

In the current literature, the map-merging problem is primarily addressed with methods based on 2D occupancy grid maps. These methods include probability~\cite{robust-multi, simultaneous-merging}, optimization~\cite{merging-occupancy, on-merging} and feature-based methods~\cite{feature-semantic}. The last ones commonly consider extracting point, line or geometric features, in order to match and merge local maps. 
Wang et al.~\cite{rtm} regarded occupancy grid maps as images and by extracting Scale-Invariant Feature Transform (SIFT) features~\cite{sift}, were able to merge local maps using the ICP scan matching algorithm~\cite{icp}. More recently, Sun et al.~\cite{subgraph} proposed a maximum common subgraph (MCS)-based occupancy grid map fusion algorithm, where the Harris corner points~\cite{harris} are extracted first, and then the maximum common subgraph is obtained using an iterative algorithm.
Finally, a transformation matrix was calculated, based on the relationship between the corner points, in order to merge the maps. Nonetheless, occupancy grid maps have limitations, especially in heterogeneous multi-robot systems. With the increased size of the explored environment, the processing power and storage, required to keep an occupancy grid map, are correspondingly significantly increasing~\cite{slam-rao}. 
In addition to this, in case of multi-modal systems, e.g. aerial and ground robots, different occupancy maps might be created, due to different operating height and viewing angles of the sensors, therefore making the accurate overlap matching process impossible. As a result, an increasing number of studies have focused on extracting features from the environment and creating feature-based 3D maps. Dense point clouds are commonly used to overcome the limitations of the occupancy maps, while are frequently used to illustrate the surroundings, as well as to extract feature points. 
The retrieved feature points are often employed in one of two scenarios: 1) inter-frame feature matching in a single robot’s local map generation process, and 2) rigid transformation computation, between local maps generated by various robots.
The extraction of stable point features, from point-feature-based maps, is an important stage in the map-merging process, since unstable point features might result in inaccurate feature matching results, jeopardizing the map-merging performance.
For such circumstances, Konolige et al.~\cite{distributed} sought to merge two local maps by matching manually extracted features (such as doors, junctions, and corners) in two local maps. Sun et al.~\cite{sparse} used the open source framework ORB\_SLAM2~\cite{orb2} to construct sparse point cloud maps by extracting FAST key points~\cite{fAST_ML} and BRIEF descriptors~\cite{brief}. The off-line dictionary’s point characteristics were then compared to those retrieved from local maps using the bag of words~\cite{bow}. Although more advanced features typically result in fewer data, the work required to extract such characteristics is also higher. Plane features~\cite{planar} are taken from the environment by fitting a plane model into numerous 3D points, and as a result, these characteristics are less vulnerable to noise, more resilient and faster to extract~\cite{point_plane}.
On the other hand, if a robot encounters a huge planar structure, it is likely that just a portion of the plane will be visible in each frame, owing to sensor FOV limitations. As a consequence, features originating from the same planar structure may show in many frames in a row and these redundant features must be merged to reflect a single planar structure in the environment. Similarly to our approach, Drwiega~\cite{3d_map_server} proposed a 3D map-merging scheme for multi-robot systems, based on overlapping regions. The overlap estimation is carried out based on the SHOT descriptors~\cite{shot} and the map alignment is performed by SAC-IA. In~\cite{3d-map-merge}, the overlap detection is made possible by first segmenting the map and then using the VFH descriptors~\cite{vfh}. Finally, the proposed KP-PDH descriptors speed up the map-merging process, offering lower computational times over the aforementioned SHOT descriptors. However, both these last two approaches could be considered as time-consuming components in an autonomous multi-robot  exploration, since the overall processing times of 15-40 seconds presented, is a substantial amount, especially for aerial platforms. Therefore, in our novel proposed approach, FRAME, we try to minimize the processing time of the map-merging algorithm by deploying deep learned descriptors, offering fast and robust result across different environments and platforms. 


\section{THE PROPOSED MAP-MERGING FRAMEWORK} \label{sec:framework}

\begin{figure*}[b]
    \centering
    \includegraphics[width=\textwidth]{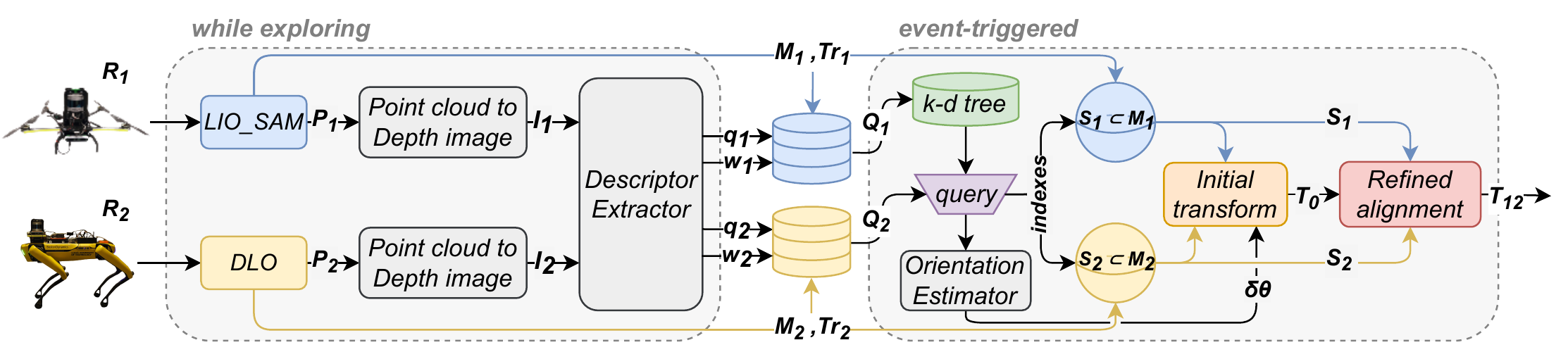}
    \caption{The overall map-merging pipeline of FRAME. While the robots $R_1, R_2$ explore the surroundings, they collect the vector sets $Q$ and $W$. A predefined event will trigger the merging process, and each robot will create its merged map $M$.}
    \label{fig:map-merge-architecture}
\end{figure*}

The scope of this research is to develop a novel 3D point cloud map-merging framework that works online, meaning that two or more robots performing an exploration mission should be able to merge their generated maps at any point, as long as there is a communication established between them and there is sufficient overlap between their maps.
Considering a system of $N$ robots $R = \{r_1, r_2, \dots, r_N \}$, in $\mathbb{R}^3$ space, each robot generates a local map $M_N$, with respect to a local static coordinate frame $\mathcal{M_N}$. The global map can be defined as a set of local maps $M = \{M^{'}_1, \dots, M^{'}_N\}$, correctly transformed to the global map frame $\mathcal{M}$. The problem can be simplified to only two inputs at a time, without loss of generality, since we can make the assumption that more than two maps can be merged recursively. In more detail, for a given time step $k$, we have two maps $M_1$ and $M_2$, which are sets of points $m \in \mathbb{R}^3$, as well as the two trajectories $Tr_1$ and $Tr_2$ that are sets of points $p \in \mathbb{R}^3$. The two point cloud maps are defined as:
\begin{equation} \label{eq:maps}
    M_1 = \{ m_{1,n_1} \}, \: M_2 = \{ m_{2,n_2} \} \:\: \text{with} \:\: n_1, n_2 \in \mathbb{N}
\end{equation}
Similarly, the trajectories are defined as:
\begin{equation}
\label{eq:trajectories}
\begin{split}
    Tr_1 &= \{p_{1,k} | k=1,2,\ldots,K\}, \\
    Tr_2 &= \{p_{2,k} | k=1,2,\ldots,K\},
\end{split}
\end{equation}
where $p_{r,k} = (x_k,y_k,z_k)_r$ represents the position of the robot $r$ at a time step $k$ relative to its local frame $\mathcal{M}_r$. The map~merging process can be described as a function $f_m$:
\begin{equation} \label{eq:functions}
    f_m : \mathbb{R}^3 \times \mathbb{R}^3 \rightarrow \mathbb{R}^3
\end{equation}
More precisely, we can denote the merged map as:
\begin{equation} \label{eq:merged}
    M = f_m(M_1,M_2) = M_1 \cup T_{12} M_2,
\end{equation}
where $T_{12}: \mathbb{R}^3 \rightarrow \mathbb{R}^3$, represents the homogeneous rigid transformation of the special Euclidean group, denoted as:
\begin{equation} \label{eq:transform}
    T_{12} = \left[ \begin{array}{cc}
         R & p\\
         0 & 1
    \end{array} \right] \in SE(3),
\end{equation}
where $R \in SO(3)$ and $p \in \mathbb{R}^3$. 
In the novel proposed approach, we divide the problem into two parts. First, our goal is to find the two overlapping submaps, yielding an initial transform $T_0 \in SE(3)$ and in the sequel we make use of the initial transform, as an initial condition to a point cloud registration algorithm, that will refine the alignment and provide the final transform $T_{12}$, as depicted in Fig. \ref{fig:map-merge-architecture}.


 \subsection{While Exploring}

Starting a mission, each robot $r$ explores the surroundings and collects information that will later be used to find the overlapping regions and then merge the maps. 
For each time step $k$, we project the point cloud $P_{r,k}$ into a spherical coordinate frame to extract a 360$^o$ of horizontal field of view, depth image~$I_{r,k}$ and then feed it to the \textit{descriptor~extractor} module that outputs the $2\times64$ vectors, $\vec q_{r,k}$ and $\vec w_{r,k}$ respectively. We define the acquired vector sets as:
\begin{gather}
    Q_r = \{\vec q \in \mathbb{R}^{64}, \, n \in \mathbb{N}: \vec q_{r,1}, \vec q_{r,2}, \dots, \vec q_{r,n} \} \\
    W_r = \{\vec w \in \mathbb{R}^{64}, \, n \in \mathbb{N}: \vec w_{r,1}, \vec w_{r,2}, \dots, \vec w_{r,n} \}
\end{gather}
More specifically, vector~$\vec q$ is an orientation-invariant, place-dependent vector, used for querying similar point clouds and $\vec w$ is an orientation-specific vector, used for regressing the yaw discrepancy between two point clouds.
These vectors are based on OREOS~\cite{oreos} and more details can be found on our extension 3DEG~\cite{3deg}. 
Along the vectors, we collect the corresponding position $p_{r,k}$ of the robot, with respect to its local map frame~$\mathcal{M}_r$.

\begin{figure}[!t] 
    \centering
    \includegraphics[width=\columnwidth]{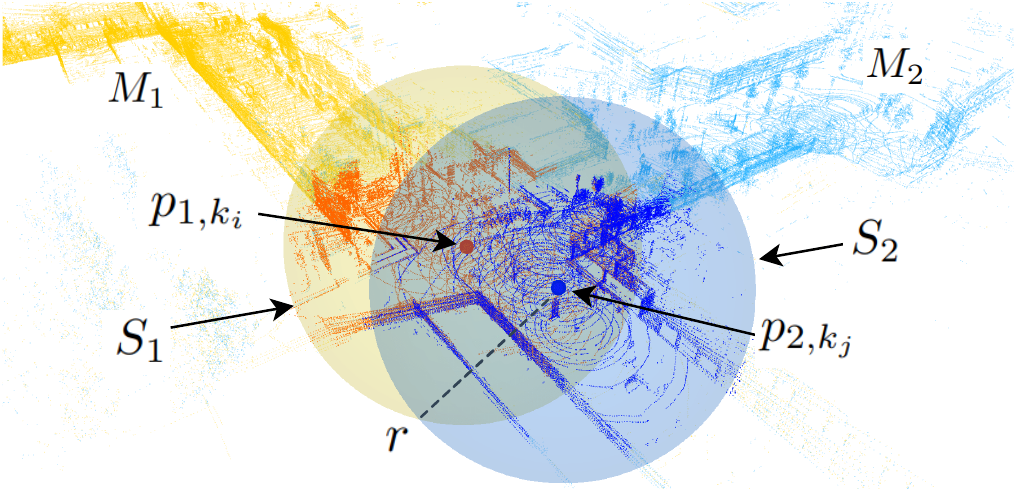}
    \caption{The merged map $M$ after the alignment of $M_1$ and $M_2$, the spheres $S_1$ and $S_2$ and their corresponding centers $p_{1,k_i}$ and $p_{2,k_j}$ of the overlapping regions. The spheres encompass the highlighted points that are used as an input to the Fast-GICP algorithm.}
    \label{fig:spheres}
    \vspace{-6.2mm}
\end{figure}
\subsection{Event-triggered}

The process of merging the maps can be triggered based on an event, depending on the given mission. For example, in a decentralized approach, the process can be initialized when there is an available connection between the two robots.
The moment the connection is established, the event will trigger the map-merging process, which is built around the following submodules.

\subsubsection{Overlap Estimation}

The first step is to utilize the aforementioned vector sets $Q_1$ and $Q_2$, in order to obtain the two overlapping regions, defined as $S_1 \subset M_1$ and $S_2 \subset M_2$. As an egocentric approach, each robot will create a \textit{k-d} tree with its own set of $Q_{1}$ vectors and query it with the incoming vector set $Q_{2}$ from the other robot. This way, we can find the pair of vectors $\vec q_{1,i}$ and $\vec q_{2,j}$ that have the minimum distance between them in the vector space and therefore were extracted from two similar point clouds.
\begin{equation}
    (k_i,k_j) = \operatorname*{arg\,min}_{(i,j) \, \in \, \mathbb{N}}f(Q_{1,i},Q_{2,j})
\end{equation}
The indices $i,j$ provide information on which time instance $k_i, k_j$ was selected from each vector set and consequently what was the position $p_{1,k_i}$ and $p_{2,k_j}$ for each robot.
This information can be used for the next step, which is determining the homogeneous initial transform $T_0 \in SE(3)$ between the past robot frames $\mathcal{R}_{1,k_i}$ and $\mathcal{R}_{2,k_j}$.

\subsubsection{Initial Transform}

To find the complete homogeneous initial transform $T_0 $, we feed the orientation-specific vectors $w_{1,k_i}$ and $w_{2,k_j}$ to the \textit{orientation estimator} module, yielding a yaw discrepancy angle prediction $\delta \theta$. The \textit{orientation estimator} is part of the descriptor extraction process and has been defined in more detail on~\cite{oreos, 3deg}.
We can then form the initial transform $T_0$ that will align the two local frames $\mathcal{R}_{1,k_i}$ and $\mathcal{R}_{2,k_j}$. 
\begin{equation}
    \label{eq:T0}
    T_{0} = \left[ \begin{array}{cc}
         R_{\hat z}(\delta \theta) & p_{1,k_i}-p_{2,k_j}\\
         0 & 1
    \end{array} \right]
\end{equation} 
Since the translation part of \eqref{eq:T0} is based on the trajectory points, we can not guarantee that the two robots explored exactly the same coordinates in the map, and therefore we can not get an exact alignment of the two point cloud maps. 


\subsubsection{Refined alignment}

To achieve the final and refined alignment, the initial transform $T_0$ is used as an initial guess in the Fast-GICP registration algorithm~\cite{fast_gicp}, providing a much faster convergence.  
To further reduce computational time, we feed into the registration algorithm only part of the two point cloud maps. Given the points $p_{1,k_i}$ and $p_{2,k_j}$ as the two centers of the overlapping regions, we sample the points inside two spherical regions $S_1$ and $S_2$ of radius $r$, defined as:
\begin{equation}
    S = \{ m, p \in \mathbb{R}^3 : ||m-p_k||^2 \leq r^2 \}
\end{equation}
As seen on Fig.~\ref{fig:spheres}, aligning the points inside the spherical regions provides the final and merged global map.
\begin{figure*}[!b]
    \begin{subfigure}{0.64\columnwidth}
    \includegraphics[width=\columnwidth]{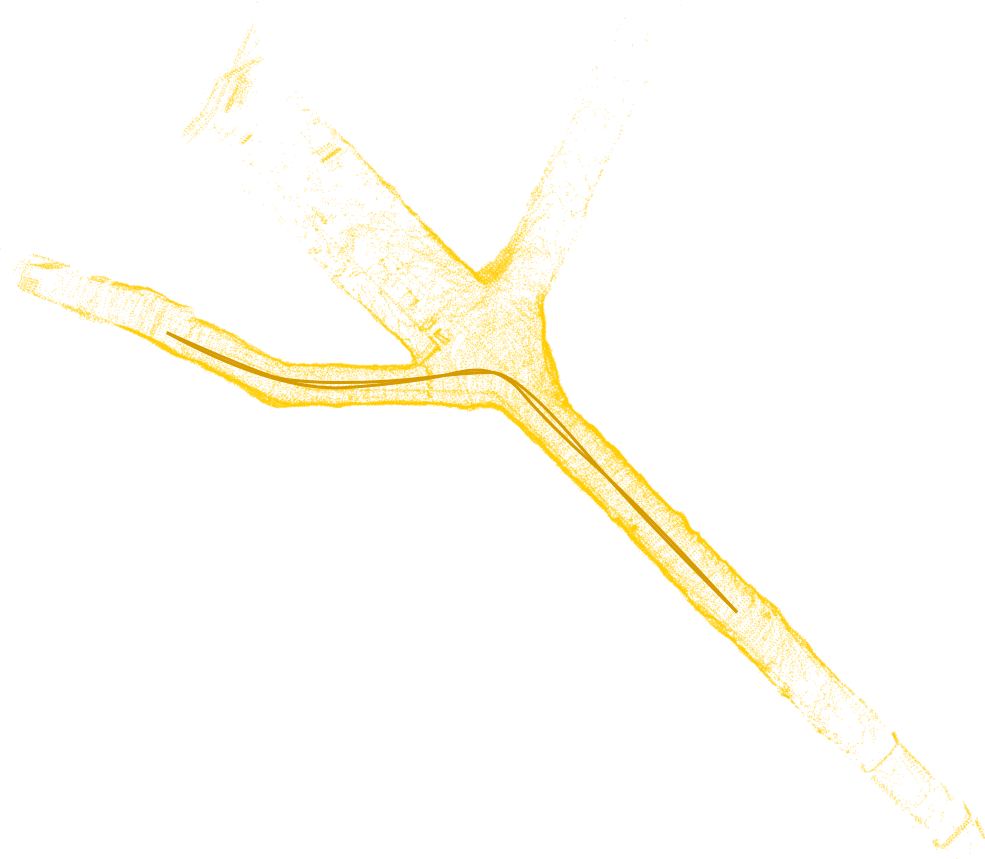}
    \caption{$M_1$ and $Tr_1$}
    \end{subfigure}
    \hfill
    \begin{subfigure}{0.44\columnwidth}
    \includegraphics[width=\columnwidth]{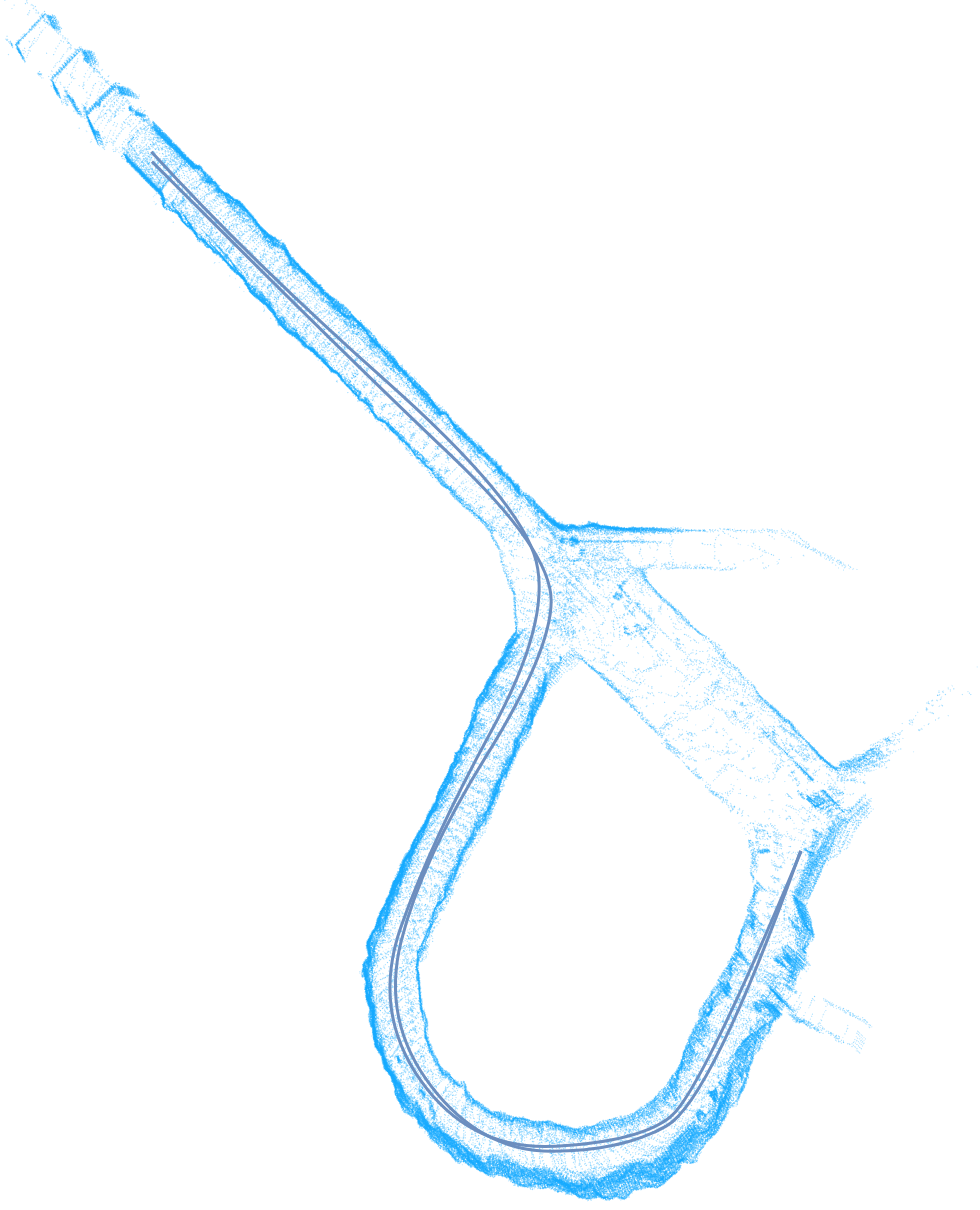}
    \caption{$M_2$ and $Tr_2$}
    \end{subfigure}
    \vspace{5pt}
    \begin{subfigure}{0.44\columnwidth}
    \includegraphics[width=\columnwidth]{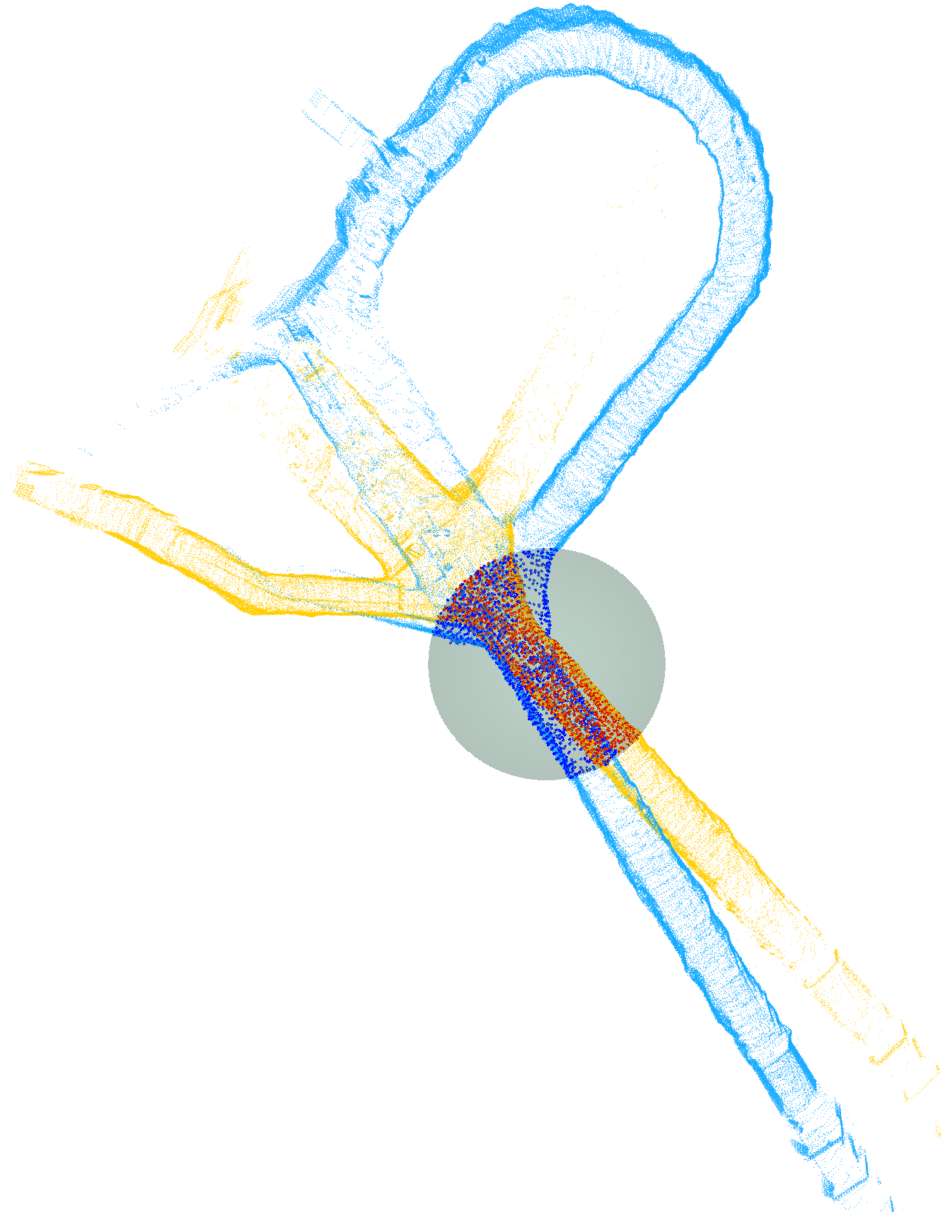} 
    \caption{Initial alignment}
    \end{subfigure}
    \hfill
    \begin{subfigure}{0.50\columnwidth}
    \includegraphics[width=\columnwidth]{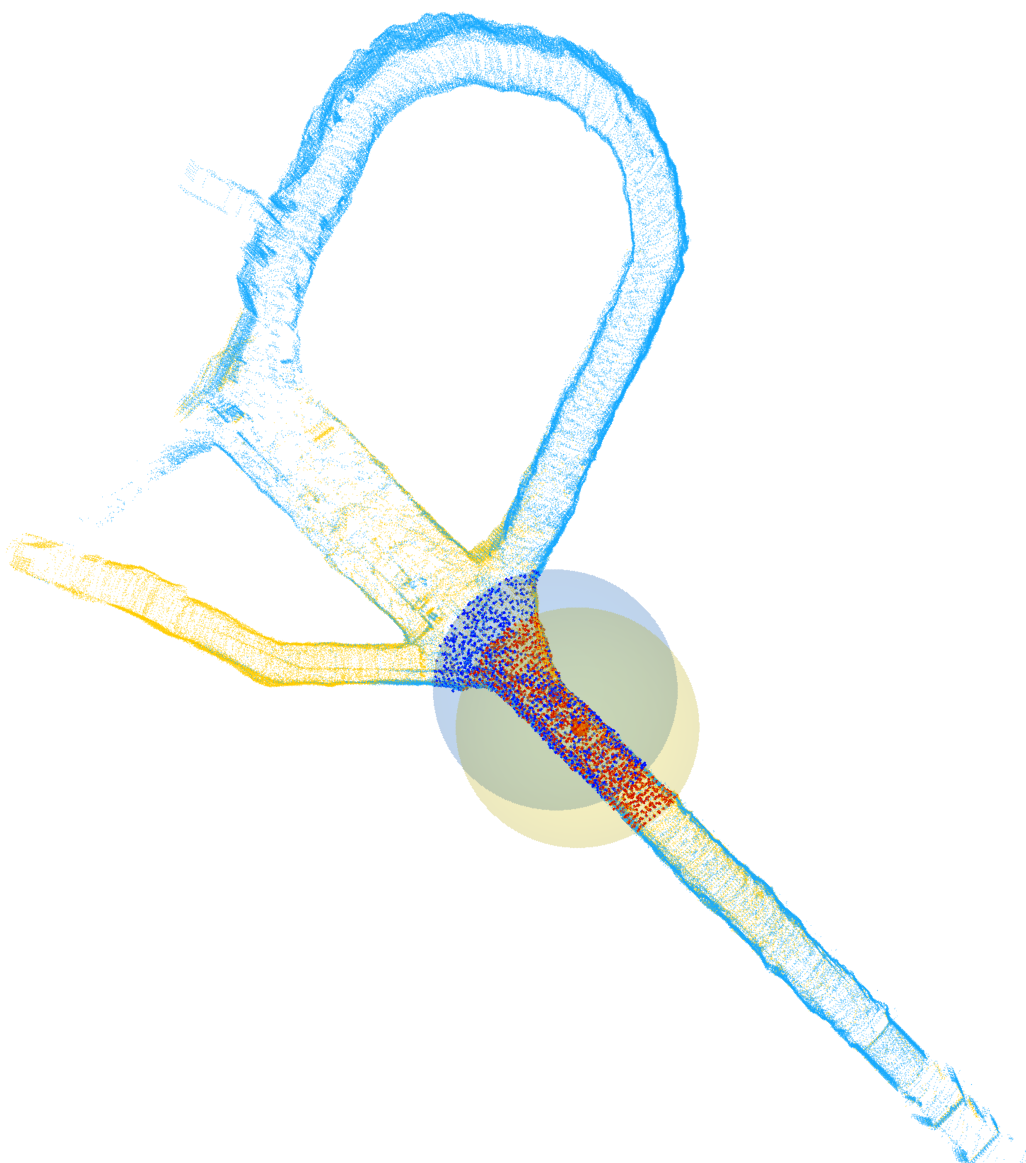}
    \caption{Refined merged map $M$}
    \end{subfigure}
    \caption{The point cloud maps $M_1$ and $M_2$ with an initial yaw difference of $180^o$, built with the aerial platform using LIO-SAM~\cite{LIO-SAM} and the legged platform using DLO~\cite{DLO} respectively. The map depicted in subfigure (c) is the initial rough alignment based on the initial transform $T_0$, and can be described as $M_0 = M_1 \cup T_{0} M_2$, while the map $M$ is the refined merged map based on the final transform $T_{12}$, defined as $M = M_1 \cup T_{12} M_2$.}
    \label{fig:mjolkberget}
\end{figure*}

\section{EXPERIMENTAL EVALUATION} \label{sec:experiments}

\subsection{Datasets and platforms}
\begin{figure*}[b!] 
    \begin{subfigure}{0.59\columnwidth}
    \includegraphics[width=\columnwidth]{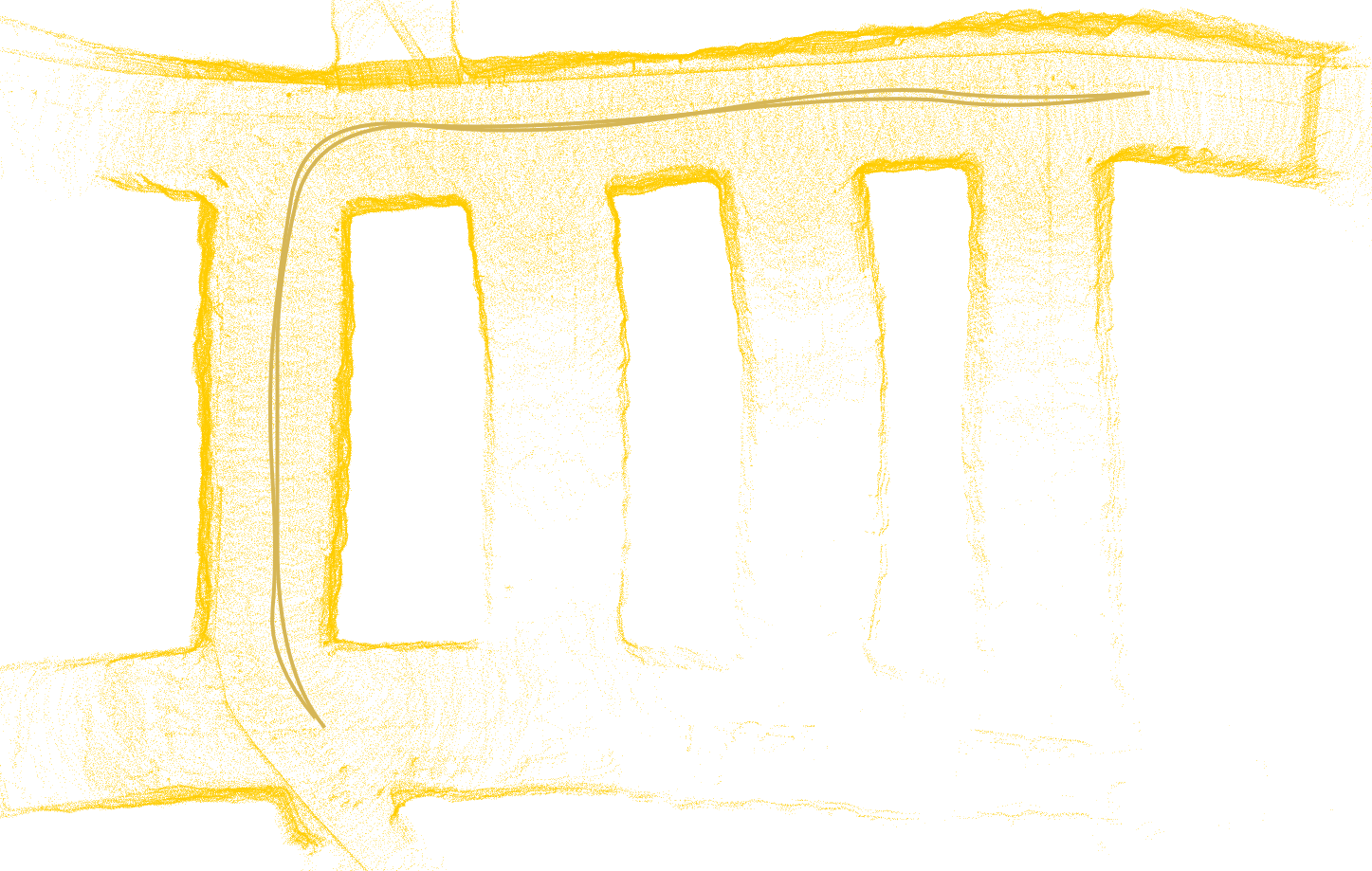} \caption{$M_1$ and $Tr_1$}
    \end{subfigure}
    \begin{subfigure}{0.60\columnwidth}
    \includegraphics[width=\columnwidth]{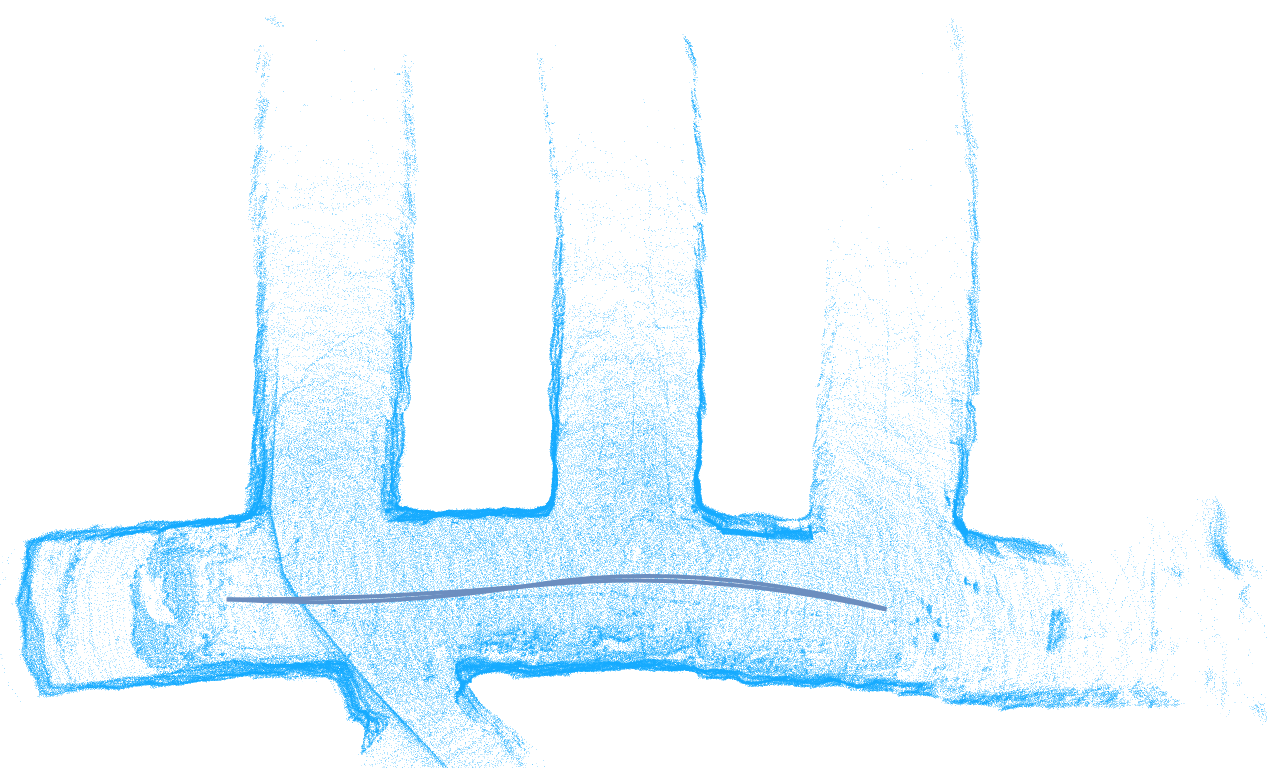} \caption{$M_2$ and $Tr_2$}
    \end{subfigure} \hfill
    \begin{subfigure}{0.34\columnwidth}
    \includegraphics[width=\columnwidth]{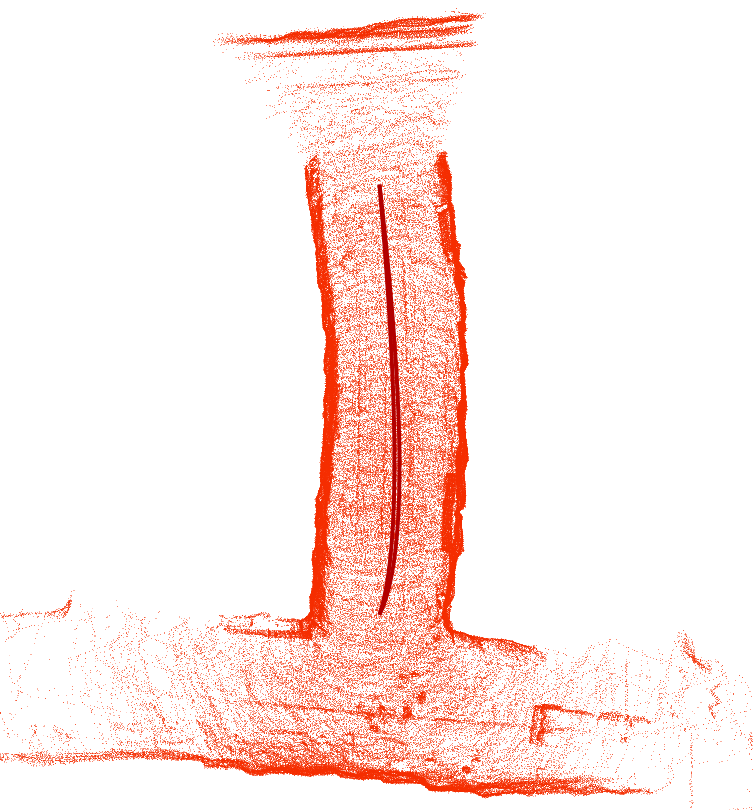} \caption{$M_3$ and $Tr_3$}
    \end{subfigure} \hfill
    \begin{subfigure}{0.37\columnwidth}
    \includegraphics[width=\columnwidth]{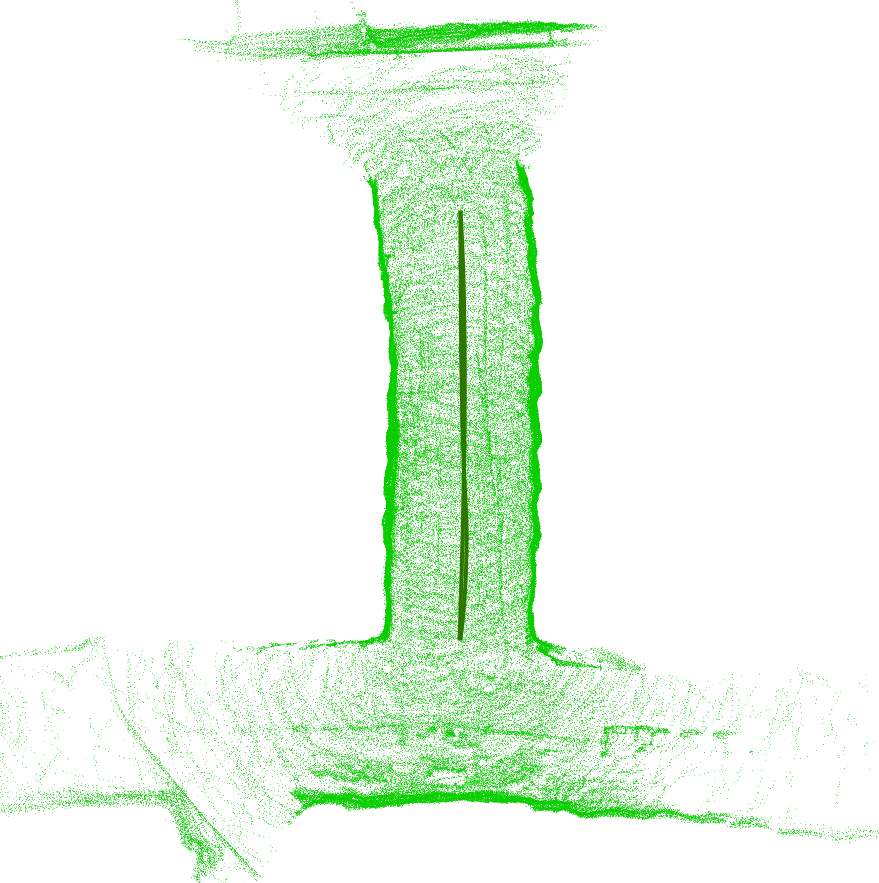} \caption{$M_4$ and $Tr_4$}
    \end{subfigure} \vspace{5pt} \\
    \begin{subfigure}{0.66\columnwidth}
    \includegraphics[width=\columnwidth]{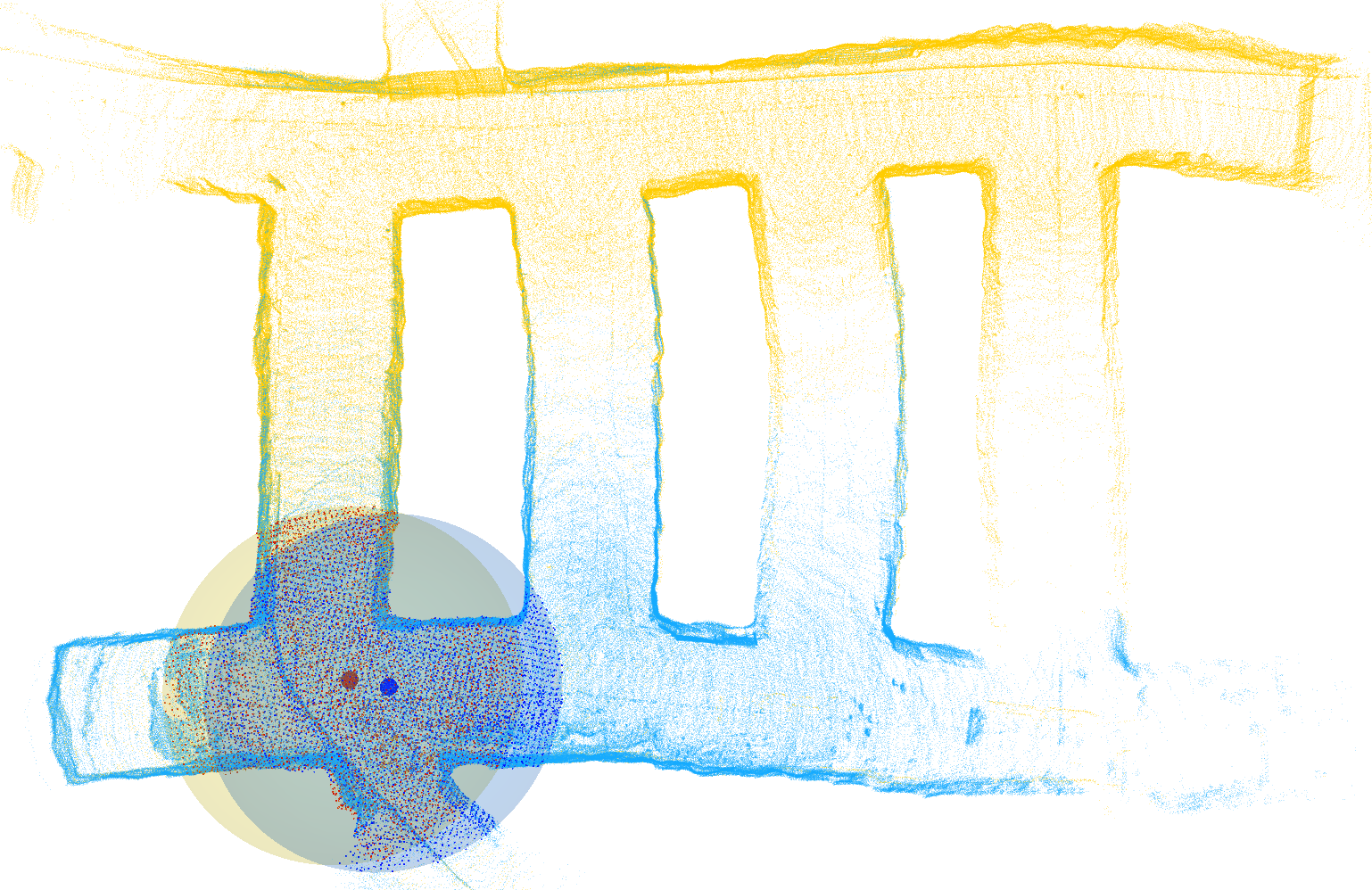} \caption{$S_1$, $S_2$ and $M_{12}$}
    \end{subfigure} \hfill
    \begin{subfigure}{0.66\columnwidth}
    \includegraphics[width=\columnwidth]{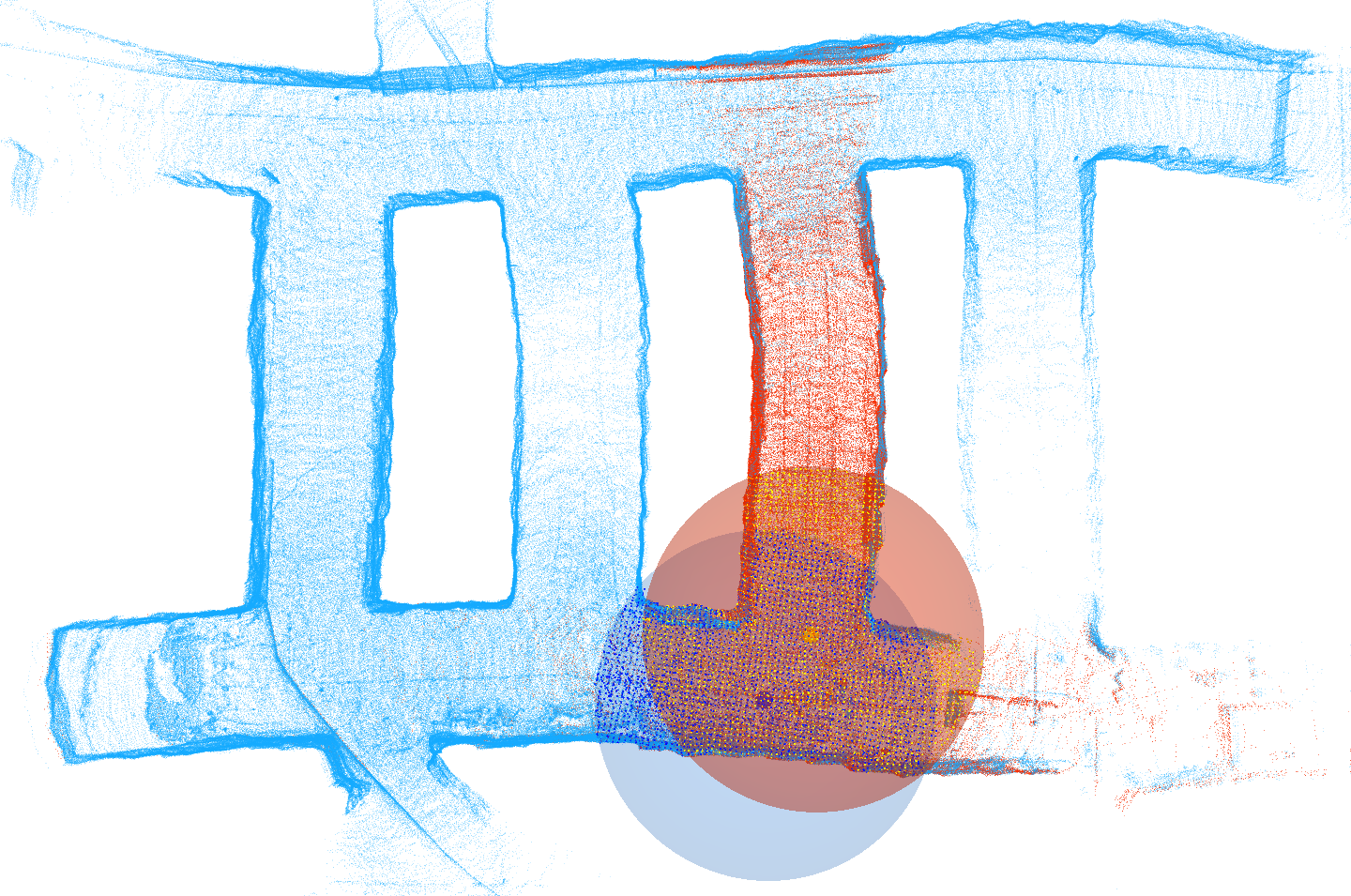} \caption{$S_{12}$, $S_3$ and $M_{13}$}
    \end{subfigure} \hfill
    \begin{subfigure}{0.66\columnwidth}
    \includegraphics[width=\columnwidth]{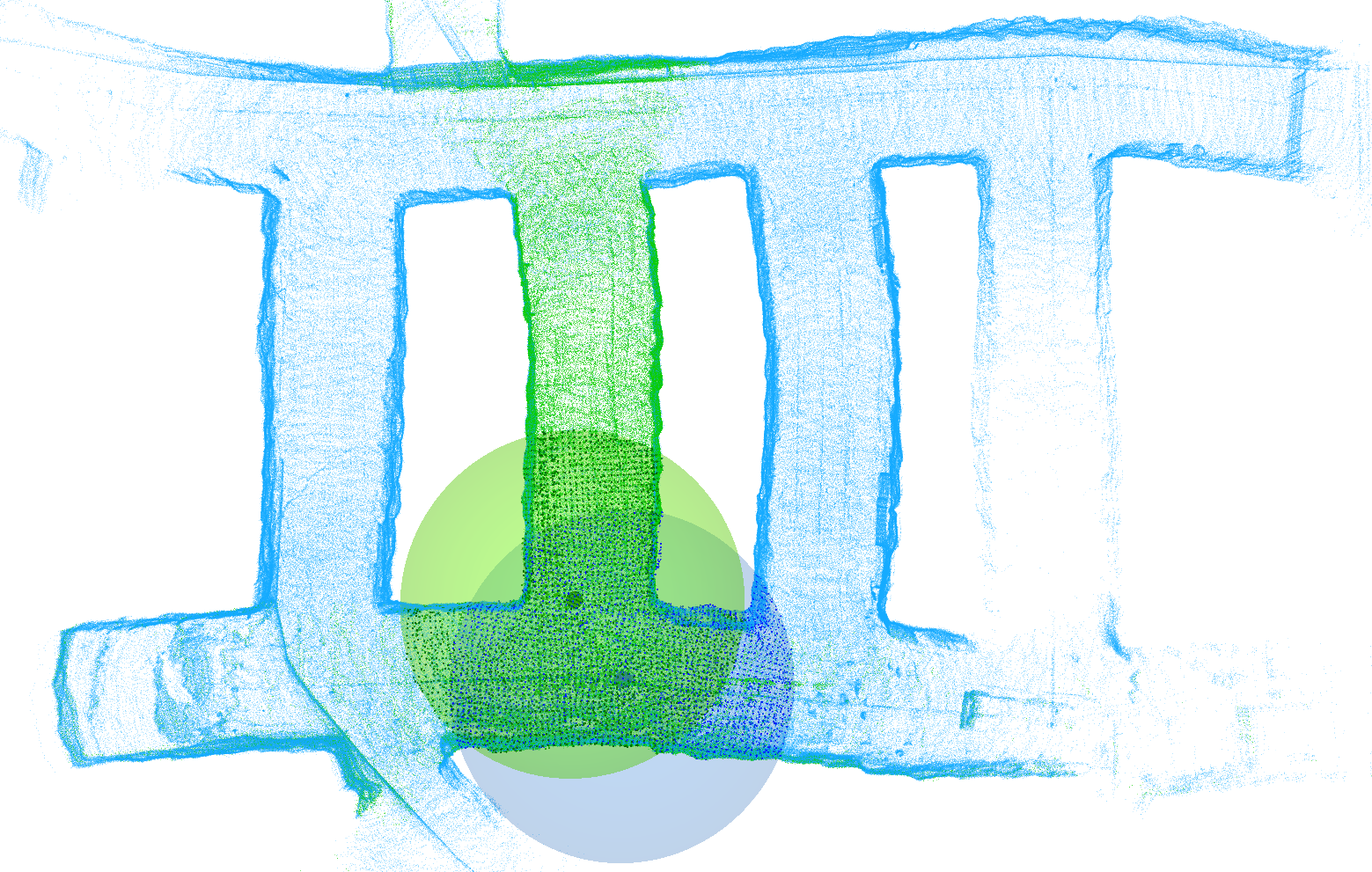} \caption{$S_{13}$, $S_4$ and $M_{14}$}
    \end{subfigure}
    \caption{A set of four maps from a real world test mine. The maps $M_1$, $M_2$, $M_3$ and $M_4$ are four individual maps that we proceed on merging sequentially. Based on the overlapping regions $S_1$ and $S_2$ with a radius of $r=15$ meters, we first get the merged map $M_{12} = M_1 \cup T_{12} M_2$, then based on $S_{12}$, $S_3$ we obtain $M_{13} = M_{12} \cup T_{13} M_3$ and finally from $S_{13}$, $S_4$ we have $M_{14} = M_{13} \cup T_{14} M_4$.}
    \label{fig:epiroc}
\end{figure*}

The proposed framework was evaluated in three different real life environments, two subterranean and one indoor.
The first environment~\cite{KOVAL2022104168} is from an underground tunnel located in Lule\r{a}, Sweden, as seen on Fig~\ref{fig:mjolkberget}.
The second one, as depicted on Fig.~\ref{fig:epiroc}, is from a real underground mine facility and unlike the first environment, it features larger tunnels, up to 10 meters wide, with multiple junctions and featureless, self-similar walls.
The last environment to test, as presented in Fig.~\ref{fig:ltu}, is the indoor corridors of one of the buildings at Lule\r{a} University of Technology, featuring long narrow and self-similar corridors.
For the experiments, two robotic platforms were used, one legged and one aerial. As a legged platform, Spot the quadruped robot from Boston Dynamics~\cite{spot} was used, and was equipped with an autonomy package~\cite{autonomy}, that includes the Velodyne Puck Hi-Res 3D LiDAR and an Intel NUC on-board computer with an Intel Core i5-10210U and 8GB of RAM.
The aerial robot is a custom-built quadrotor~\cite{compra} carrying a Velodyne VLP16 PuckLite 3D LiDAR and the same on-board computer as the legged robot.
The main difference between the two LiDARs is that even though both feature 16 channels, the Hi-Res has $20^o$ of vertical field of view, while the PuckLite has $30^o$.
This directly affects the quality and similarity of the depth images produced by each robot, and the learned descriptors extracted in the early stages of the framework's pipeline. Finally, the integration algorithms are implemented using the ROS framework, both on Ubuntu 18.04 and Melodic version, as well as Ubuntu 20.04 and Noetic version.

\begin{table*}[t!] 
\centering
\vspace{5mm}
\setlength{\belowcaptionskip}{-6pt}
\caption{The experimental metric results, where $M_1$ and $M_2$ represent the amount of points in each point cloud.} \label{table:results}
\begin{adjustbox}{width=\textwidth}
\begin{tabular}{ccccccccc} 
\toprule
 & $M_1$ & $M_2$ & $Tr_1$ (m) & $Tr_2$ (m) & OVERLAP (\%) & $T_e$ (m) & $R_e$ (deg) & TIME (s) \\ 
\toprule
Fig.~\ref{fig:mjolkberget} (a)-(b) & $1.59 \cdot 10^5$ & $1.61\cdot 10^5$ & 136 & 257 & 33 & 0.092 & 3.495 & 0.229 \\
Fig.~\ref{fig:epiroc} (a)-(b) & $5.14\cdot 10^5$ & $5.95\cdot 10^5$ & 113 & 139 & 35 & 0.161 & 1.031 & 0.325 \\
Fig.~\ref{fig:epiroc} (a)-(b)-(c) & $1.11\cdot 10^6$ & $2.82\cdot 10^5$ & 252 & 86 & 82 & 0.119 & 0.286 & 0.360 \\
Fig.~\ref{fig:epiroc} (a)-(b)-(c)-(d) & $1.39\cdot 10^6$ & $2.73\cdot 10^5$ & 338 & 161 & 87 & 0.079 & 0.229 & 0.391 \\
Fig.~\ref{fig:ltu} (a)-(b) & $8.29\cdot 10^5$ & $1.28\cdot 10^6$ & 843 & 530 & 35 & 0.175 & 2.234 & 0.427 \\
\toprule
\end{tabular}
\end{adjustbox}
\vspace{-5mm}
\end{table*}
\subsection{Metrics}
 
To evaluate the performance of FRAME against existing methods, we introduced the metrics that are shown in Table~\ref{table:results}. In this comparison with state-of-the-art approaches, we present the number of points for each set of maps, the traveled distance for each trajectory, the approximate overlap percentage between the two maps, the transform error and the computational time. Since we are testing in indoor or underground environments, GPS data are not available to provide a ground truth. For that, we make use of the CloudCompare~\cite{cloud_compare}, an open source software that provides point cloud alignment and merging, while by importing two point cloud maps, roughly aligning them by hand and then letting the software refine them, we extract the final translation $T_{gt}$ and final rotation $R_{gt}$, which will be used as the ground truth. The translation error $T_e$ and rotation error $R_e$ are defined as:
\begin{equation}
    T_e = ||T_{gt} - T||, \:\: R_e = ||R_{gt} R^{-1} - I_3||,
\end{equation}
where $T$ is the translation part and $R$ is the rotation matrix of the final transform $T_{12}$.
For the evaluation process, we compare our framework against two other publicly available methods, map-merge-3D~\cite{map-merge-3d} and 3D map server~\cite{3d_map_server}.
For the rest of this article, we will refer to them as MM3D and 3DMS, respectively.
The first algorithm initially pre-processes the point cloud maps to remove outliers, then performs a 3D feature extraction algorithm with SIFT points or Harris corners and finally compares features to find correspondences and align the two point clouds.
The second method that was highlighted in the Section~\ref{sec:related_work}, relies on the SHOT descriptors for the overlap estimation and on the SAC-IA for the alignment.


\subsection{Results}

After extensive tuning, to the best of the author's capability, with the parameters and the different kind of descriptors each package offers, we were unable to get a satisfying result in any of the maps presented below. 
We start the experimental evaluation with the maps $M_1$ and $M_2$ of Fig.~\ref{fig:mjolkberget}, that have approximately $150 \cdot 10^3$ points after traveling a total distance of $136$ and $257$ meters, respectively. The map $M_1$ is made on the aerial platform running LIO-SAM~\cite{LIO-SAM} while the map $M_2$ is made with the legged robot running DLO~\cite{DLO}. The first package, MM3D, was unable to regress the 180$^o$ yaw difference between the two maps, therefore resulting in a high rotational error $R_e$, as seen in Table~\ref{table:simplified}. 
Even though the second package, 3DMS, was able to estimate the yaw difference $\delta \theta$, it incorrectly matched the two side corridors and resulted in a high roll angle error. 
To further evaluate these packages against the proposed framework, we simplify the first pair of maps. 
The maps $M_1$ and $M_2$ of Fig.~\ref{fig:mjolkberget} are built with different platforms, sensors and SLAM algorithms, leading to different point densities. 
For that, we rebuild map $M_2$ with the same platform and configuration as map $M_1$ and we intentionally reduce the yaw discrepancy from $180^o$ to approximately 15$^o$. 
As one can observe on Table~\ref{table:simplified}, this time around MM3D is able to yield a sufficient result but with a greater computational time of approximately 9 seconds compared to FRAME that has a higher translation accuracy in just under 0.3 seconds. 
The 3DMS is still unable to output a good result after 260 seconds of computational time.

\begin{table}[t!] 
\centering
\vspace{1.5mm}
\caption{The experimental metric results for the original set of maps from Fig.~\ref{fig:mjolkberget} and for the simplified version.} \label{table:simplified}
\begin{adjustbox}{width=\columnwidth}
\begin{tabular}{ccccccc} 
\toprule
 & \multicolumn{3}{c}{ORIGINAL } & \multicolumn{3}{c}{SIMPLIFIED} \\ 
\toprule
 & T$_e$ (m) & R$_e$ (deg) & TIME (s) & T$_e$ (m) & R$_e$ (deg) & TIME (s) \\ 
\toprule
MM3D~\cite{map-merge-3d} & 70.78 & 159.52 & 30.78 & 1.13 & 3.56 & 9.21 \\
3DMS~\cite{3d_map_server} & 5.43 & 110.42 & 191.39 & 9.70 & 16.52 & 260.47 \\
FRAME &\textbf{ 0.09} & \textbf{3.43} & \textbf{0.25} & \textbf{0.09} & \textbf{3.43} & \textbf{0.27} \\
\toprule
\end{tabular}
\end{adjustbox}
\vspace{-6mm}
\end{table}
\begin{figure*}[b!]
    \vspace{-6mm}
    \begin{subfigure}{0.33\textwidth}
    \includegraphics[width=\textwidth]{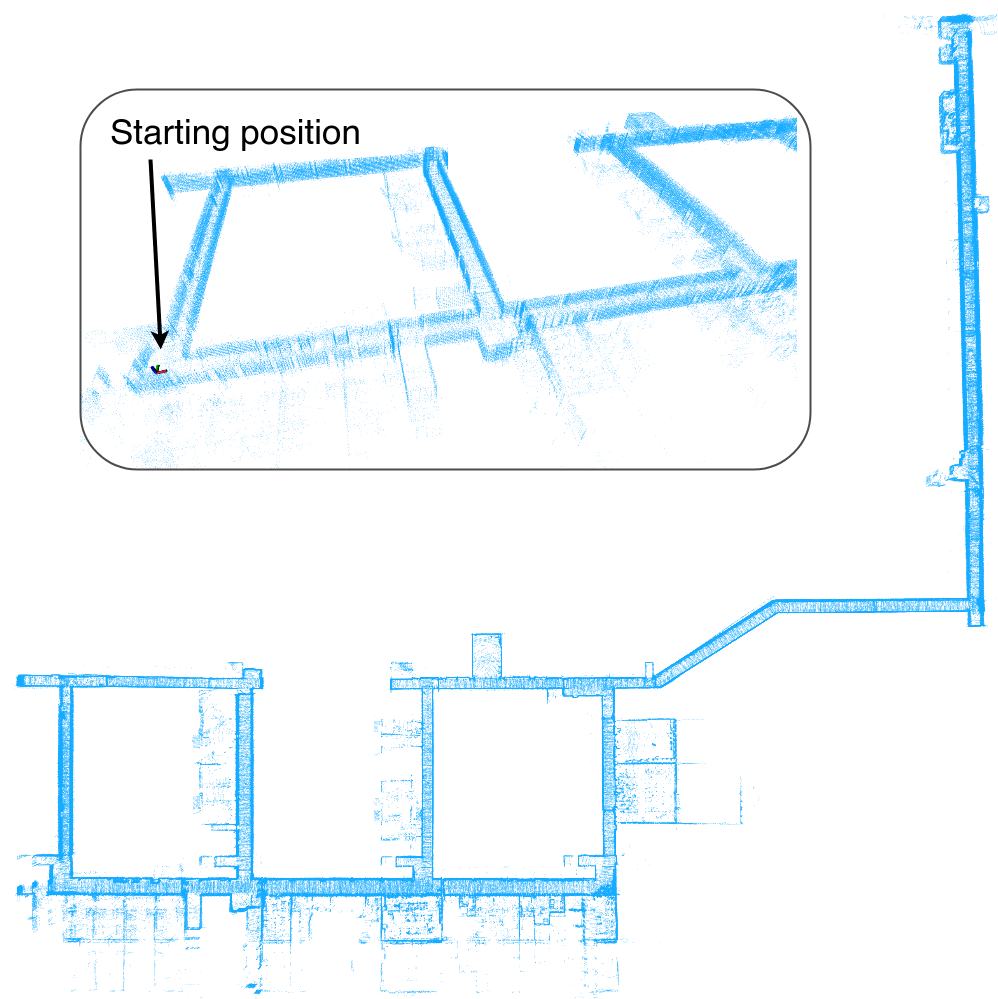} \caption{Map $M_1$}
    \end{subfigure}
    \begin{subfigure}{0.34\textwidth}
    \includegraphics[width=\textwidth]{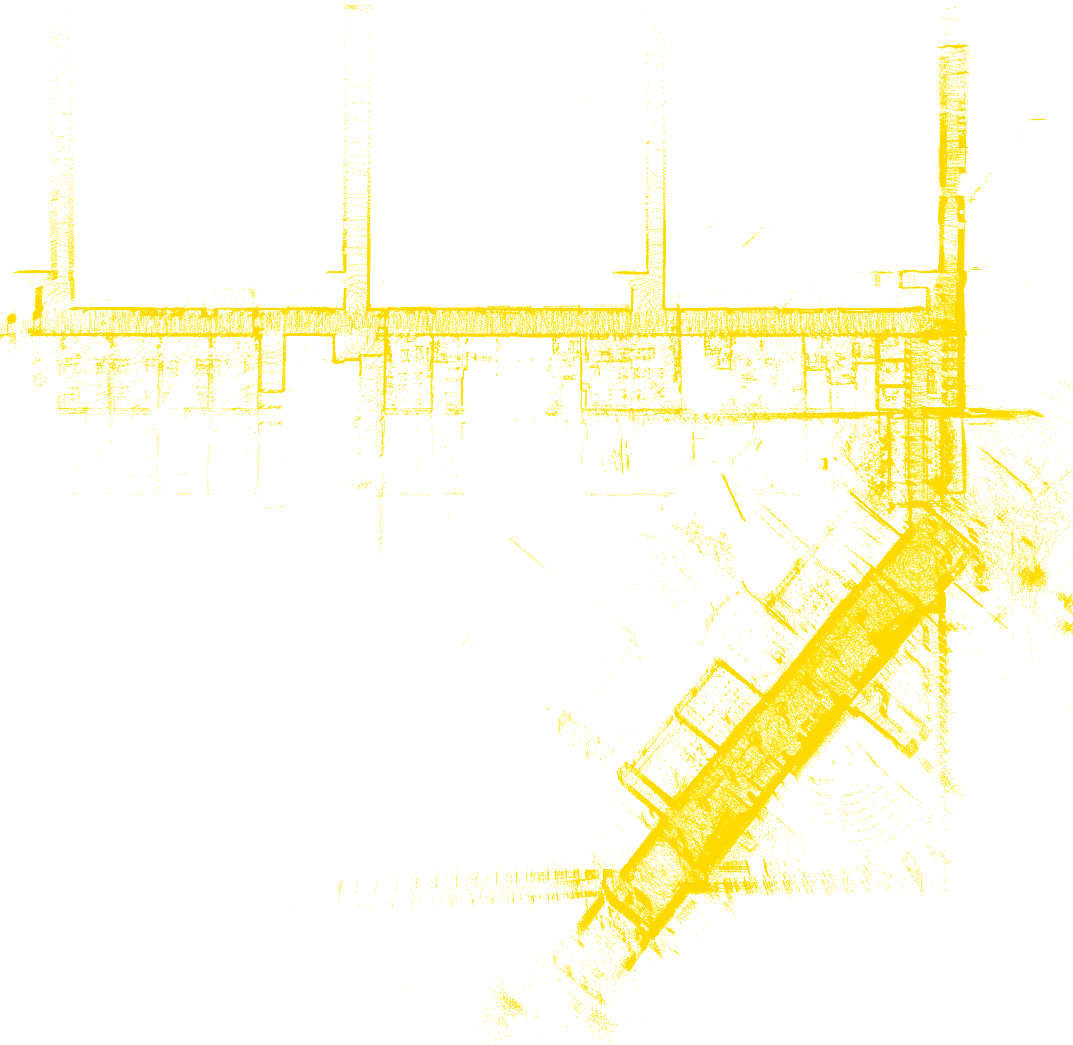} \caption{Map $M_2$}
    \end{subfigure}
    \begin{subfigure}{0.32\textwidth}
    \includegraphics[width=\textwidth]{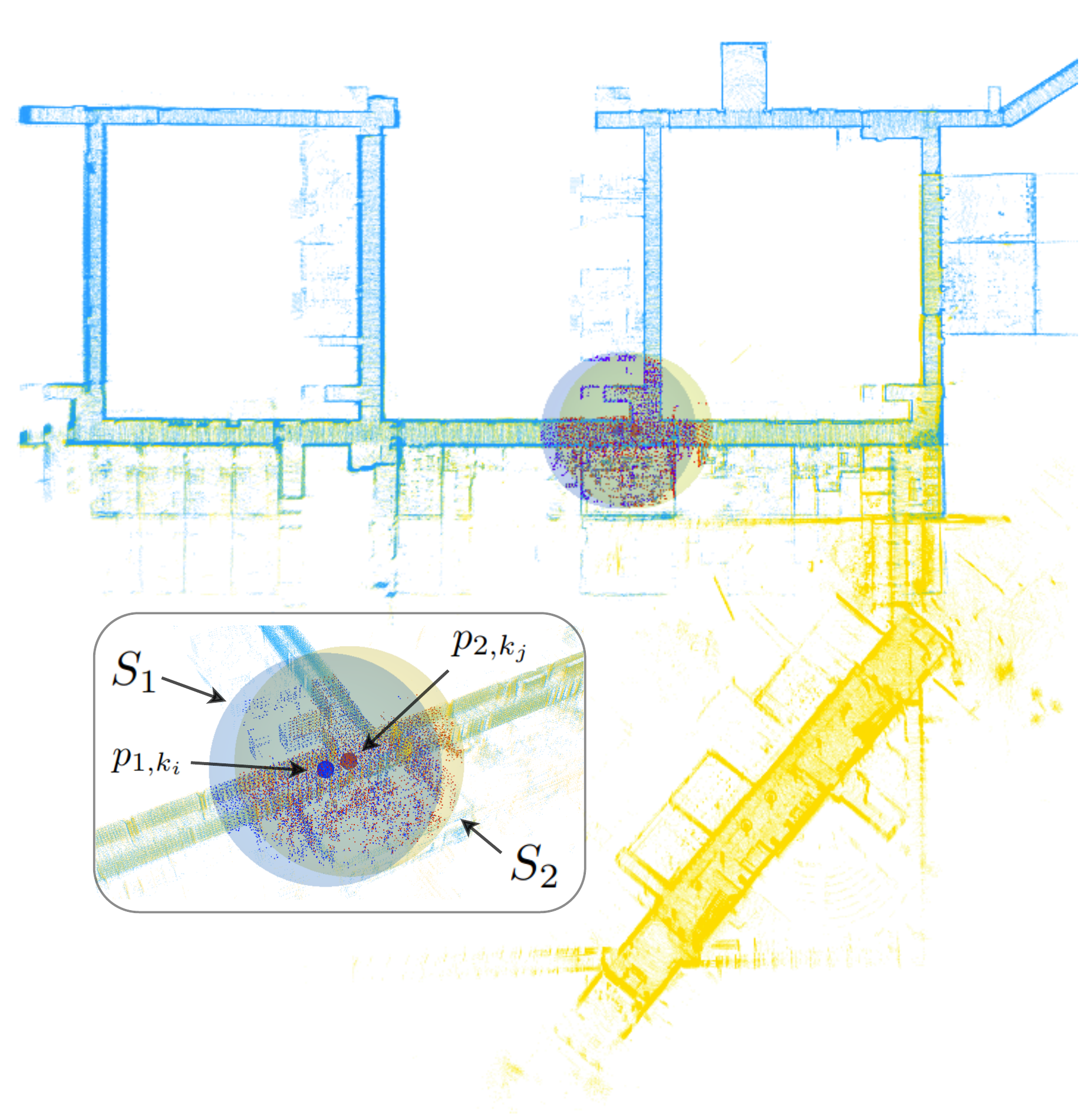} \caption{Merged map $M$ and spheres $S_1$, $S_2$}
    \end{subfigure}
    \caption{Larger scale indoor environment from Lule\r{a} University of Technology, where the two robots start together but explore different branches of the building.}
    \label{fig:ltu}
\end{figure*}

As a sequel, we consider the subterranean case of Fig.~\ref{fig:epiroc}, with four instances $M_1$, $M_2$, $M_3$ and $M_4$.
In this scenario, only the aerial platform was used to produce the maps consisted of $500-600 \cdot 10^3$ points for $M_1$ and $M_2$ and $280 \cdot 10^3$ points for $M_3$ and $M_4$.
Initially, we merge the maps $M_1$ and $M_2$ yielding the transform $T_{12}$ and the merged map $M_{12} = M_1 \cup T_{12} M_2$ and continue the process by using the merged map $M_{12}$ in the place of $M_1$ and repeating the same steps as many times as needed.
Same as before, both the aforementioned packages failed to compute a correct transform. The initial yaw difference was approximately 90$^o$ and both packages rotated map $M_2$ in the opposite direction, trying to match the two longer sections of each map.
Maps $M_3$ and $M_4$ are particularly challenging as they are not only similar to each other but to the first corridor connecting $M_1$ and $M_2$ as well. Due to the lack of distinct features for the descriptors to rely on, the algorithms we compare against, match both maps $M_3$ and $M_4$ to the first corridor, as the higher point density provides a better matching score. 
On the other hand, FRAME manages to detect the overlapping regions $S_{12}$, $S_3$ and $S_3$, $S_{13}$, even though the trajectories $Tr_3$ and $Tr_4$ from each robot did not overlap with $Tr_2$.  By providing a good initial guess $T_0$ to the registration algorithm, adjusting the sphere radius to $r=15$ meters, to encompass for the large openings, and by keeping the correspondence threshold radius low, we are able to align the point clouds keeping the translation $T_e$ and rotation $R_e$ error low.

The final part of the experiments was carried out in the basement corridors of Lule\r{a} University of Technology and consists of two long runs of approximately 800 and 500 meters traveled distance, using the legged platform. The set of maps $M_1$ and $M_2$, depicted on Fig.~\ref{fig:ltu}, consist of $800 \cdot 10^3$ and $1 \cdot 10^6$ points per map and include long featureless corridors that both the MM3D and 3DMS were unable to distinguish and align properly. On the contrary, the proposed framework is able to identify one of the intersections as the overlapping region set $S_1$ and $S_2$, with a radius of $r=10$~meters, and proceed on first making an initial guess $T_0$ based on the trajectory points $p_{1,k_i}$, $p_{2,k_j}$ and then refining the alignment in a total amount of computational time of less than half a second, highlighting the benefit of querying learned place recognition descriptors $\vec{q}$ and $\vec{w}$.


\section{CONCLUSIONS} \label{sec:conclusions}

In this article, we have presented and successfully evaluated a novel event-triggered, egocentric map-merging framework for 3D point cloud maps, referred to as FRAME, that can be directly implemented in multi-robot exploration. 
The utilization of learned place recognition descriptors, as well as yaw discrepancy regression descriptors, enabled to have a fast querying process that reduces computational time as it replaces the global descriptor extraction and feature matching step, traditionally used in the map-merging solutions. 
Furthermore, the derived initial transform, provides a good initial guess to the registration algorithm and therefore we have a fast convergence with low translational and rotational errors.
Finally, through our experiments we have successfully demonstrated fast and robust performance of the proposed framework that promotes robotic autonomy for longer missions that include not only multiple but multi-modal robots, with different sensor configurations.



\addtolength{\textheight}{-2.7cm}   





\bibliographystyle{./IEEEtranBST/IEEEtran}
\bibliography{./IEEEtranBST/IEEEabrv,root}

\end{document}